\documentclass[sigconf]{acmart}

\usepackage{amsfonts}
\usepackage{amsmath}
\usepackage{mathrsfs}
\usepackage{subfigure}
\usepackage{enumitem}
\usepackage[switch]{lineno}
\usepackage{multirow}
\usepackage{graphicx}
\usepackage{makecell}
\usepackage{balance}
\AtBeginDocument{%
  \providecommand\BibTeX{{%
    \normalfont B\kern-0.5em{\scshape i\kern-0.25em b}\kern-0.8em\TeX}}}

\copyrightyear{2021} 
\acmYear{2021} 
\setcopyright{acmlicensed}\acmConference[MM '21]{Proceedings of the 29th ACM International Conference on Multimedia}{October 20--24, 2021}{Virtual Event, China}
\acmBooktitle{Proceedings of the 29th ACM International Conference on Multimedia (MM '21), October 20--24, 2021, Virtual Event, China}
\acmPrice{15.00}
\acmDOI{10.1145/3474085.3475547}
\acmISBN{978-1-4503-8651-7/21/10}

\acmSubmissionID{2026}

\begin{document}
\fancyhead{}
\title{Exploring the Quality of GAN Generated Images for Person Re-Identification}


\author{Yiqi Jiang}
\authornote{Both authors contributed equally to this research.}
\email{yiqi.jyq@alibaba-inc.com}

\author{Weihua Chen}
\authornotemark[1]
\email{kugang.cwh@alibaba-inc.com}

\affiliation{%
	\institution{Alibaba Group}
	\country{China}
}

\author{Xiuyu Sun}
\authornote{Corresponding author}
\email{xiuyu.sxy@alibaba-inc.com}
\author{Xiaoyu Shi}
\email{linyin.sxy@alibaba-inc.com}

\affiliation{%
	\institution{Alibaba Group}
	\country{China}
}

\author{Fan Wang}
\email{fan.w@alibaba-inc.com}
\author{Hao Li}
\email{lihao.lh@alibaba-inc.com}

\affiliation{%
  \institution{Alibaba Group}
  \country{China}
}

\begin{abstract}
Recently, GAN based method has demonstrated strong effectiveness in generating augmentation data for person re-identification (ReID), on account of its ability to bridge the gap between domains and enrich the data variety in feature space. However, most of the ReID works pick all the GAN generated data as additional training samples or evaluate the quality of GAN generation at the entire data set level, ignoring the image-level essential feature of data in ReID task. In this paper, we analyze the in-depth characteristics of ReID sample and solve the problem of ``What makes a GAN-generated image good for ReID''. Specifically, we propose to examine each data sample with id-consistency and diversity constraints by mapping image onto different spaces. With a metric-based sampling method, we demonstrate that not every GAN-generated data is beneficial for augmentation. Models trained with data filtered by our quality evaluation outperform those trained with the full augmentation set by a large margin. Extensive experiments show the effectiveness of our method on both supervised ReID task and unsupervised domain adaptation ReID task.
\end{abstract}

%



\maketitle

\section{Introduction}
\label{sec:intro}

Given a query image, person re-identification (ReID) is the task of retrieving person images of the same identity from the gallery consisting of images collected across multiple cameras. Cameras often differ from each other regarding resolution, viewpoints, and illumination, which result in drastic changes in appearance and background of person images. 
Therefore, it has been crucial for ReID to learn representations that are robust against intra-class variations.
Many methods have been proposed to learn a stable feature representation among different cameras, such as KISSME~\cite{koestinger2012large} and DNS~\cite{zhang2016learning}. 
Relying on the strong representation power of convolutional neural networks (CNNs), ReID methods~\cite{zheng2016person,sun2018beyond} achieve more powerful deep embeddings through deep learning. But it is still very challenging to learn a robust and discriminative representation to handle the appearance variance across domains (\textit{e.g.} different cameras or different datasets).

\begin{figure}[t]
\begin{center}
\includegraphics[width=0.99\linewidth]{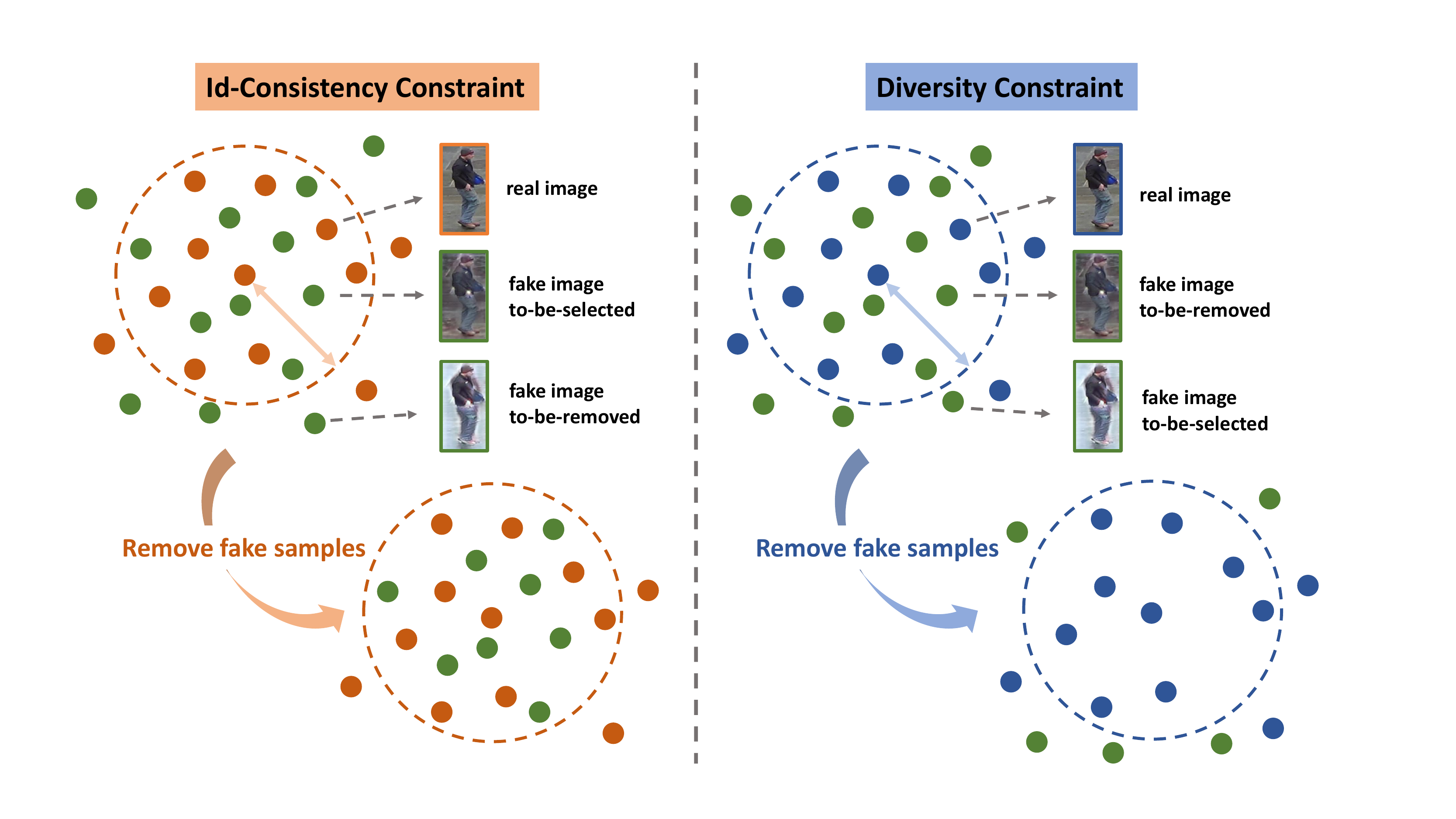}
\end{center}
   \caption{The illustration of id-consistency constraint (left) and diversity constraint (right). Id-consistency constraint favors the generated images which are close to the corresponding real ones in consistency space. Diversity constraint discards the generated images which are within a certain distance with the real image in diversity space.}
   \Description{top and bottom is the illustration of id-consistency constraint and diversity constraint, respectively. Id-consistency constraint favors the generated images which are close to the corresponding real ones in consistency space. Diversity constraint discards the generated images which are within a certain distance with the real image in diversity space.}
\label{fig.1}
\end{figure}

Recently, more works exploit generative adversarial network (GAN)~\cite{goodfellow2014generative} to introduce additional augmented data into training set~\cite{zheng2017unlabeled} so that ReID model can “see” more intra-class variations during training. For example, CamStyle~\cite{zhong2018camera} uses CycleGAN~\cite{zhu2017unpaired} to transfer images from one camera to the style of other cameras, and PTGAN~\cite{wei2018person} applies human pose conditioned GANs to generate pedestrian images of the same identity but with different poses. 
Focusing on domain adaptive person ReID, SPGAN~\cite{deng2018image}, CR-GAN~\cite{chen2019instance} and PDA-Net~\cite{li2019cross} transfer labeled images from source domain into target domain to learn discriminative models on target domain. 
Although existing generative methods can synthesize plenty of visually pleasing data, there is no guarantee that all the generated images are beneficial to the final ReID training. 

In this paper, We introduce two constraints when evaluating whether the generated images are favorable for ReID augmentation.
\begin{itemize}
\item \textbf{Id-consistency constraint}. The generated images should preserve identity information consistent with the real ones as much as possible.
\item \textbf{Diversity constraint}. The generated images should diverse from the real images as much as possible to introduce more variations.
\end{itemize}
To better describe these constraints, we map the generated samples onto two different feature spaces, consistency space and diversity space. Intuitively, the consistency feature space represents identity-related appearance information in an image (\textit{e.g.} clothing, hair, gender), which is the major information used for identifying a particular person. The diversity feature space contains all variations which are not shared within the images of the same identity (\textit{e.g.} pose, illumination, camera views). As illustrated in Figure \ref{fig.1}, an ideal generated image for augmentation, should be close to the corresponding real image in consistency space, and also should keep a certain distance from the real image in diversity space. An image that fails to satisfy these two constraints would be discarded because it might bring negative impact during training.
Therefore, we design a metric-based sampling method to select a subset from the full set of augmentation data. Our experiments show that training with the filtered augmentations could achieve better ReID results compared with that trained with the full set of generated data.

Our contributions can be summarized as below:
\begin{itemize}
\item[\textbf{1)}]
We demonstrate that GAN generated images are not all beneficial for ReID training. Consistency and diversity constraints are presented to assess whether a generated image is suitable for ReID data augmentation.

\item[\textbf{2)}]
The principles of consistency and diversity feature space are provided. Example projections for each feature space are presented, which we believe are great choice to be applied in practice.

\item[\textbf{3)}]
Experiments show that, by mapping generated images to consistency and diversity feature spaces and filtering images with a simple sampling method, the augmentation set becomes more beneficial for ReID training, and the resulting ReID model achieves better performance compared with the model trained with full set of augmentation data.
\end{itemize}


\section{Related Works}
\label{sec:relatedwork}
\subsection{Deep Learning-based ReID Methods}

Recent methods mainly rely on CNNs for its strong representation power to learn metric spaces and discriminative feature representations to handle data variations. 
Zheng~\emph{et al.}~\cite{zheng2016person} regards ReID problem as a classification problem, considering each person as a particular class. Wu~\emph{et al.}~\cite{wu2018and} and Zheng~\emph{et al.}~\cite{zheng2017discriminatively} combine identification loss with verification loss to achieve better metric space. Cheng~\emph{et al.}~\cite{cheng2016person},~Hermans~\emph{et al.}~\cite{hermans2017defense}, and Ristani~\emph{et al.}~\cite{ristani2018features} apply triplet loss with hard sample mining and achieve greater improvement in performance. Several recent works~\cite{lin2019improving,su2016deep,wang2018transferable} employ pedestrian attributes and multi-task learning to enrich feature learning with more supervisions. Alternatives employ pedestrian alignment and human part matching to leverage on the human structure prior. Li~\emph{et al.}~\cite{li2017person} and Sun~\emph{et al.}~\cite{sun2018beyond} split input images or feature maps horizontally to take advantage of local spatial cues. Other work, like ResNet-IBN~\cite{pan2018two}, improves feature representation by enhancing backbone with instance normalization~\cite{huang2017arbitrary} to better catch style migration.

\subsection{GAN-based Augmentation Methods}

CNN training might suffer from under-fitting if training set doesn't possess enough variation or there is cross-domain problem where we don't have any labeled data in target domain. With recent progress in generative adversarial networks (GANs)~\cite{goodfellow2014generative}, generative models have become an appealing choice to solve these problem. Zheng~\emph{et al.}~\cite{zheng2017unlabeled} use DC-GAN~\cite{radford2015unsupervised} to improve the discrimination ability of learned CNN embeddings. Qian~\emph{et al.}~\cite{qian2018pose} and Wei~\emph{et al.}~\cite{wei2018person} enrich the training space by generating different pose images. Zhong~\emph{et al.}~\cite{zhong2018camera} propose CamStyle data augmentation approach which transfers images from one camera to the style of another camera. FD-GAN~\cite{ge2018fd} generates a new person image of the same identity as the input with a given pose. Different from the works of Wei~\emph{et al.}~\cite{wei2018person} and Zhong~\emph{et al.}~\cite{zhong2018camera}, it distills identity-related and pose-unrelated features from the input image, getting rid of pose-related information disturbing the ReID task.



\subsection{Quantitative Measurements for GAN}

Measuring the quality of GAN generated images has attracted extensive attention in various applications. One of the most common ways to evaluate GANs is the Inception Score~\cite{salimans2016improved}. It uses an Inceptionv3 network~\cite{szegedy2015going} pretrained on ImageNet to compute the logits of the generated samples. Similar to the Inception Score, Fréchet Inception Distance~\cite{heusel2017gans} also relies on Inception’s evaluation to measure quality of generated samples. Different from evaluating GAN results on image level, FID takes the features from the Inception’s penultimate layer and estimates Gaussian approximations from both real and generated images, measuring quality by computing distribution distance. GANtrain and GANtest~\cite{shmelkov2018good} take generated data and real data as training sample to train a classification network and evaluate on the other set, respectively. Scores of generated images are given by comparing the performance of GANtrain and GANtest with a baseline network trained and evaluated both on real data. FID and GANtrain/GANtest evaluate GAN quality on set-level, which estimate whether the distribution of generated images is good or not, while the problem remains untouched that whether they are suitable to serve as augmentation data for training. In this paper, we focus on assessing the GAN generated image at image level for their potential benefits in training a ReID model.


\begin{figure*}
\begin{center}
\includegraphics[width=0.99\textwidth]{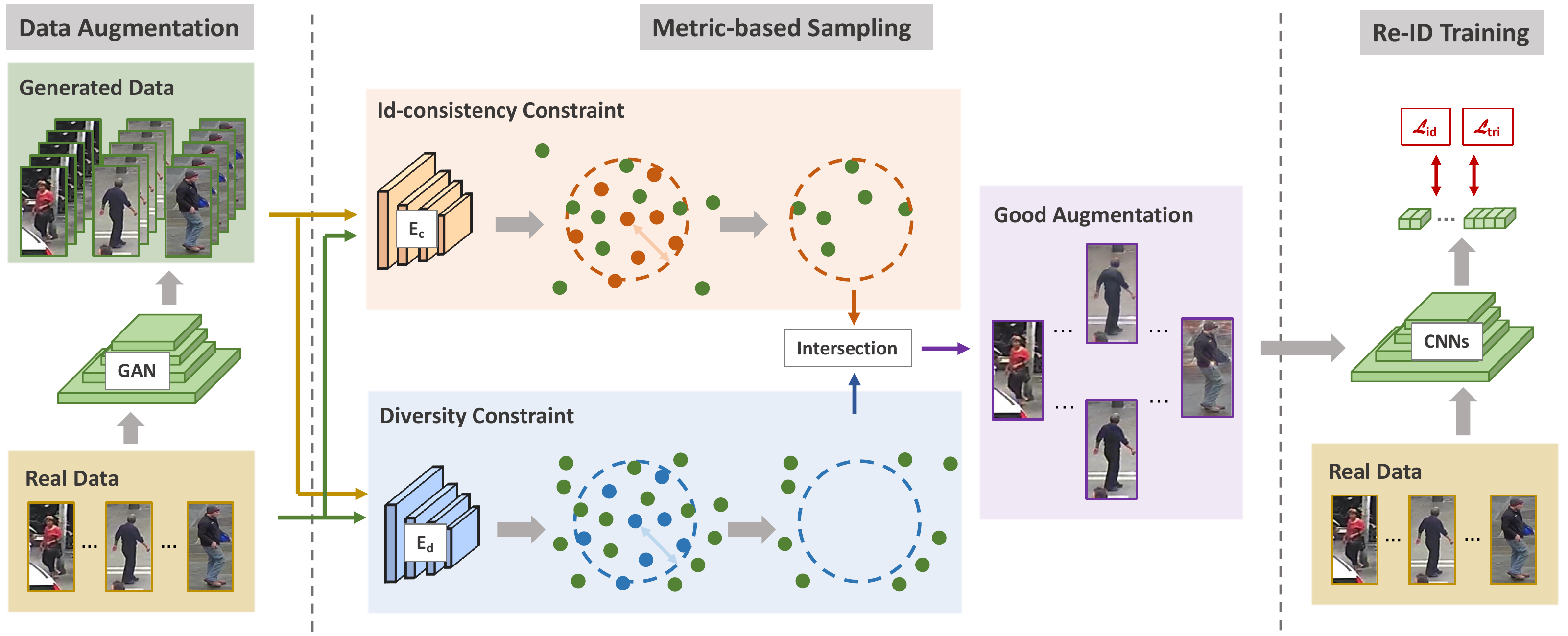}
\end{center}
   \caption{The pipeline of the proposed method. For each real image, we firstly use GAN to generate more images for augmentation. Then, Consistency encoder $E_{c}$ and diversity encoder $E_{d}$ are applied to both real data and generated data to extract consistency features and diversity features. Compared with the corresponding real image, generated images with small distance in consistency space and large distance in diversity space are selected, respectively. The intersection of images in consistency space and images in diversity space is applied to achieve good augmentations. Finally, good augmentations are combined with real data to train the ReID task.}
   \Description{The pipeline of the proposed method. For each real image, we firstly use GAN to generate more images for augmentation. Then, Consistency encoder $E_{c}$ and diversity encoder $E_{d}$ are applied to both real data and generated data to extract consistency features and diversity features. Compared with the corresponding real image, generated images with small distance in consistency space and large distance in diversity space are selected, respectively. The intersection of images in consistency space and images in diversity space is applied to achieve good augmentations. Finally, good augmentations are combined with real data to train the ReID task.}
\label{fig.2}
\end{figure*}

\section{The Proposed Method}
\label{sec:proposedmethod}
The pipeline of our proposed method is illustrated in Figure \ref{fig.2},
Firstly, GAN models are applied to generate more images for augmentation. Then, $E_{c}$ and $E_{d}$ is provided as projectors to map both real and generated images onto consistency and diversity feature space, respectively. Metric-based sampling method is further applied to select good augmentations. Finally, good augmentations are combined with real data to train the ReID task. Details of above modules are introduced as follows. 

\subsection{Projection Ecoders}
\label{ssec:projections}

We introduce two constraints, id-consistency constraint and diversity constraint, to assess the suitability of GAN generated image for augmentation purpose. In practice, the generated images are mapped onto two different feature space to estimate whether they satisfy these constraints. Two encoders are provided as feature space projectors and are denoted as $E_c$ and $E_d$, respectively.

 
\noindent\textbf{Consistency Feature Encoder $E_c$} The consistency feature space represents identity related information, such as face, body shape, clothing, hair, gender, and other attributes which do not change across cameras and could help identify a particular person. Ideally, a ReID model could represent images in a feature space which is id-consistent in the domain of training data. The better the model being trained, the more compact the feature distribution would be for each identity cluster.
 
Furthermore, some researchers present representation disentangling methods~\cite{lee2018diverse,ge2018fd,eom2019learning,zheng2019joint} which decompose feature into identity related and unrelated explicitly to enhance the representation power in id-related space. Inspired by this line of research, we build the encoder $E_c$ based on the identity related branch in ISGAN~\cite{eom2019learning}.
 
Specifically, to disentangle id-related and unrelated features, two encoders $E_R$ and $E_U$ join the structure of GAN. Given a pair of images $I_a$ and $I_p$ of the same identity, $E_R$ and $E_U$ extract identity-related features $\phi_R(I_a)$ and $\phi_R(I_p)$, and identity-unrelated features, $\phi_U(I_a)$ and $\phi_U(I_p)$. After shuffling and regrouping $\phi_R$ and $\phi_U$, the GAN reconstructs images and force the reconstructed image to recreate the original input when $I_a = I_p$, and to generate images with same identity when $I_a \neq I_p$. The shuffling loss and identity loss is defined as following:

\begin{equation}
    L_S = \sum_{i,j \in {a,p}}\left\|I_i-G(\phi_R(I_j) \oplus \phi_U(I_i))\right\|_1 
\end{equation}

where $\oplus$ represents the concatenation of features, please refer to \cite{eom2019learning} for more details.
With this constraint design, the encoder $E_R$ extracts feature with a strong identity consistency, which makes it eligible to serve as our consistency space encoder $E_c$.

\noindent\textbf{Diversity Feature Encoder $E_d$} Diversity feature space contains all feature variations which are not shared within images of the same identity (\textit{e.g.} poses, lighting conditions, resolutions, viewpoints). These features enrich the variance in ReID samples and could also bridge the domain gap between different scenarios or datasets.

Generally, there are many candidate models which could be regarded as an encoder or a projector to map images to diversity feature space, but to different extent. A random initialized model projects images to a random feature space, which could capture random representations of diversities in ReID. The id-unrelated encoder branch of disentangling model ISGAN, seems to be another good choice, but most of the constraints in ISGAN are focus on enhancing the id-related encoder $E_R$, leaving constraint on unrelated encoder $E_U$ a Gaussian noise constraint.

ImageNet~\cite{Russakovsky2015ImageNet} is a large scale classification data set and contains a great deal of variations in illuminations, view point, \emph{etc}. A well-trained classification model on ImageNet possesses powerful representation ability, which makes it able to build a rich diversity feature space. As a result, we choose a ResNet-50~\cite{2016Deep} pretrained on ImageNet as one of our diversity space encoders.

\subsection{Metric-based Sampling}
\label{ssec:sampling}

As mentioned above, two encoders, $E_c$ and $E_d$, are built to extract consistency features and diversity features, and map each image onto these two feature spaces simultaneously. In each space, we calculate the id center $C_i$ of all real image features for every identity cluster $i$, denoted as $C_c^i$ in consistency space and $C_d^i$ in diversity space.
\begin{gather}
    C_c^i=\frac{1}{n_i} \sum_{j=1}^{n_i}E_c(x_j), \qquad C_d^i=\frac{1}{n_i} \sum_{j=1}^{n_i}E_d(x_j),
\end{gather}
where $n_i$ denotes the number of images within identity $i$, and $x_j$ represents the $j_{th}$ image in identity $i$. 

For each generated image with identity $i$, L2 distance between its feature and the corresponding id center is computed in the two feature spaces:
\begin{gather}
    d_c^j=\left\|E_c(x_j) - C_c^i\right\|_2, \qquad d_d^j=\left\|E_d(x_j) - C_d^i\right\|_2,
\end{gather}

 A simple sampling rule is designed based on the distance to measure the effectiveness of features in each space.
Specifically, to narrow down the domain gap in consistency feature space, we keep the generated images whose distance to $C_c^i$ are less than a specified threshold $T_c$ and regard these images as consistency-candidates. To enrich potential variations in diversity feature space, the generated images whose distances to $C_d^i$ are greater than a specified threshold $T_d$ are kept as diversity-candidates.
A sample beneficial to ReID training should satisfy above requirements in both consistency and diversity space. Therefore, an intersection of both candidate sets is made to generate sampled augmentation data set.
\begin{equation}
    S_{sampled} = S_{d_c<T_c} \cap S_{d_d>T_d}
\end{equation}
where $S_{d_c<T_c}$ and $S_{d_d>T_d}$ denote consistency candidates and diversity candidates respectively.

Besides the criteria of filtering images one by one, their relationship should also be taken into consideration. Generated images which are close to each other in the diversity feature space do not bring in much diversity information; instead, too many duplicated training samples would increase training time and even cause imbalanced training which needs to be treated carefully.

To handle this problem, we employ Local Outlier Factor (LOF)~\cite{breunig2000lof} to monitor the density of each generated image in diversity feature space. If the image holds a high density, we will randomly drop it with a probability of $\alpha$, which is set to 0.3 in our experiments by experience.

The final sampled augmentation can be obtained by
\begin{equation}
    S_{final} = S_{d_c<T_c} \cap S_{d_d>T_d} \cap S_{lof}
\label{equ:sfinal}
\end{equation}
where $S_{lof}$ indicates the diversity candidate images filtered by LOF-based monitor in diversity feature space.

\subsection{ReID Training}

Given the new training set consists of all the real images and the sampled fake images, the training strategy should be carefully designed to exploit the effectiveness of these augmentations.

Firstly, it's important to control the balance between real and fake images during training. We use $B=P \times K$ images to form the mini-batch, where $P$ and $K$ denotes the number of different person-ids and the number of different images per person-id, respectively. In $K$, we set $K=M+N$, where $M$ and $N$ indicates the number of real and fake training samples. By setting the ratio of $\frac{M}{N}$, we can control the effect of real and fake images on training. 

Secondly, most of works combine cross-entropy loss and triplet loss together to train ReID model. However, from experiments, we found the triplet loss on the fake data brings a negative effect to the model. A possible explanation is that there are still noises in the fake augmented data to some extent. The triplet loss is more sensitive to the noise than the classification loss, because a noise sample would affect all the related pairs in the triplet loss.
As a result, we only use the cross-entropy loss for training on fake augmented images. 

The loss function is described as:
\begin{equation}
    L_{ReID} = \frac{1}{N} \sum_{i=1}^{N}L_R^i + \frac{1}{M} \sum_{i=1}^{M}L_F^i,
\label{equ:sampling}
\end{equation}
where $L_R^i= L_{id}(x_R^i) + L_{tri}(x_R^i)$ and $L_F^i=L_{id}(x_F^i)$. $x_R^i$ is the real image and $x_F^i$ is the generated image. $L_{id}$ is cross-entropy loss function and $L_{tri}$ is triplet loss function.

Thirdly, along the assumption above, the fake augmented data still contain noise to some extent. We involve Label Smoothing Regularization(LSR)~\cite{li2017person} to alleviate the impact of noise. Specifically, we apply LSR on both the real images and the fake images to softly distributed their labels as follow:
\begin{equation}
q_{LSR}(c)=
\begin{cases}
1 - \epsilon + \frac{\epsilon}{C}& c = y \\
\frac{\epsilon}{C}& c \neq y,
\end{cases}
\end{equation}
where $\epsilon$ is a small constant to encourage the model to be less confident on the training set. As fake images contain much more noise, in our study, $\epsilon_{r}$ is set to be 0.1 for real images and $\epsilon_{f}$ is set to be 0.3 for fake images, respectively. 

\begin{table*}[t]
\caption{Comparisons with the baseline model and state-of-the-art supervised ReID methods on Market-1501, DukeMTMC-ReID and MSMT-17 datasets (\%). The best and second best results are presented in \textbf{bold} and \underline{underline}.}
\label{table:SupSOTA}
\resizebox{0.92\textwidth}{!}{
\begin{tabular}{|l|p{1.8cm}<{\centering}|p{1cm}<{\centering}|p{1cm}<{\centering}|p{1cm}<{\centering}|p{1cm}<{\centering}|p{1cm}<{\centering}|p{1cm}<{\centering}|}
\hline
\multicolumn{2}{|c|}{\multirow{2}{*}{Method}} & \multicolumn{2}{c|}{Market-1501} & \multicolumn{2}{c|}{DukeMTMC-reID} & \multicolumn{2}{c|}{MSMT-17} \\
\cline{3-8}
\multicolumn{2}{|c|}{} & Rank@1 & mAP & Rank@1  & mAP & Rank@1 & mAP \\
\hline
\hline
MGN \cite{MGN} & MM2018 & 95.7 & 86.9 & 88.7 & 78.4 & 76.9 & 52.1 \\ 
ISGAN \cite{eom2019learning} & NIPS2019 & 95.2 & 87.1 & \underline{90.9} & 79.5 & - & - \\
DGNet \cite{zheng2019joint} & CVPR2019 & 94.8 & 86.0 & 86.6 & 74.8 & - & - \\
PGFA \cite{miao2019pose} & ICCV2019 & 91.2 & 76.8 & 82.6 & 65.5 & - & - \\
OSNet\cite{zhou2019omni} & ICCV2019 & 94.8 & 84.9 & 88.6 & 73.5 & 78.7 & 52.9 \\
CBN \cite{camera-bn} & ECCV2020 & 91.3 & 77.3 & 82.5 & 67.3 & 72.8 & 42.9 \\
SAN \cite{SAN} & AAAI2020 & \underline{96.1} & 88.0 & 87.9 & 75.7 & 79.2 & 55.7 \\
SCSN \cite{SCSN} & CVPR2020 & 95.7 & 88.5 & \bf{91.0} & 79.2 & \bf{83.8} & 53.5 \\
HOReID \cite{wang2020high} & CVPR2020 & 94.2 & 84.9 & 86.9 & 75.6 & - & - \\
ISP \cite{ISP} & ECCV2020 & 95.3 & \underline{88.6} & 89.6 & \bf{80.0} & - & - \\
\hline
Baseline & - & 95.1 & 87.8 & 89.6 & 79.1 & 81.1 & \underline{56.5} \\
OriginalCamStyle & - & 94.9 & 85.5 & 86.6 & 75.6 & 80.6 & 53.7  \\
ModifiedCamStyle & - & 95.5 & 87.9 & 88.4 & 78.2 & 81.0 & 56.2 \\
{\bf Ours} & - & {\bf 96.2} & {\bf 89.2} & {\bf 91.0} & \underline{79.9} & \underline{81.9} & {\bf 57.2} \\
\hline
\end{tabular}}
\end{table*}

\begin{figure*}
\begin{center}
\includegraphics[width=0.91\textwidth]{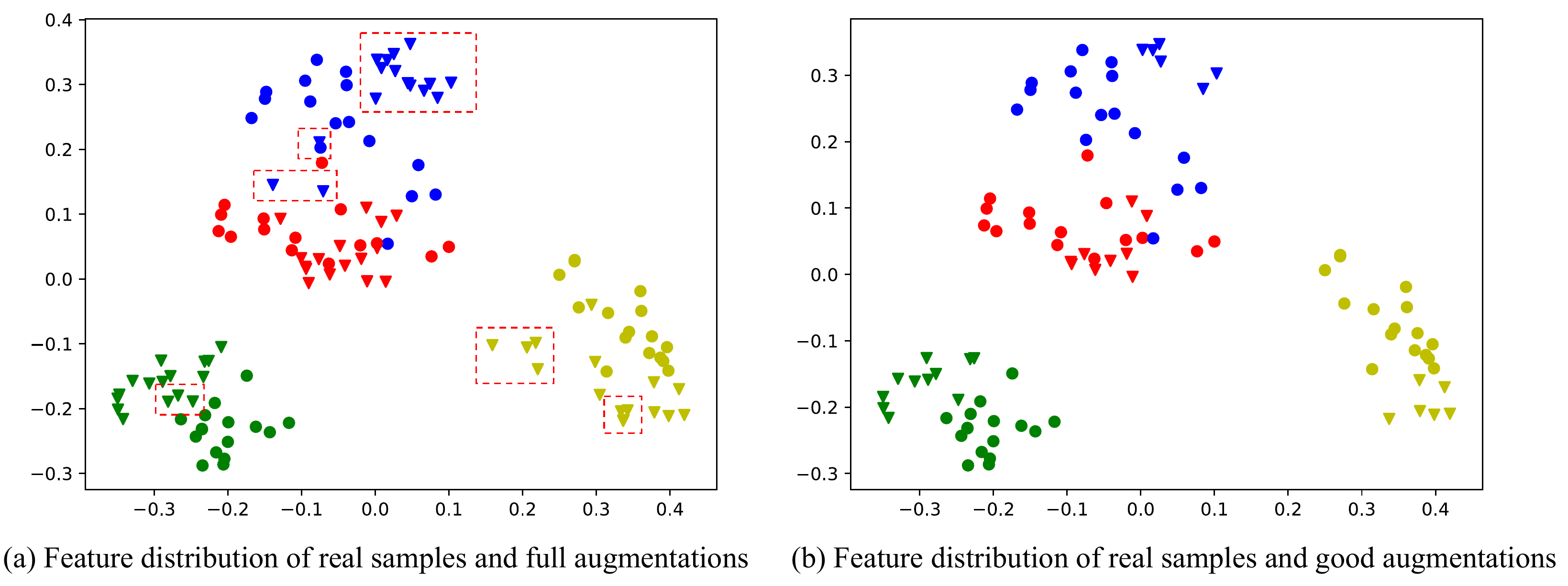}
\end{center}
  \caption{t-SNE~\cite{maaten2008visualizing} visualization on Market-1501. Circle dots denote the real samples and triangles denote the generated samples. Different colors represent different identities. (a) The feature distribution of real samples and full augmentation. (b) The feature distribution of real samples and good augmentations. Full augmentations introduce a large number of noisy or duplicate samples (red boxes in (a)). By proper sampling based on the proposed consistency constraint and diversity constraint, most of bad samples are removed, as shown in (b).}
  \Description{t-SNE~\cite{maaten2008visualizing} visualization on Market-1501. Circle dots denote the real samples and triangles denote the generated samples. Different colors represent different identities. (a) The feature distribution of real samples and full augmentation. (b) The feature distribution of real samples and good augmentations. Full augmentations introduce a large number of noisy or duplicate samples (red boxes in (a)). By proper sampling based on the proposed consistency constraint and diversity constraint, most of bad samples are removed, as shown in (b).}
\label{fig.3}
\end{figure*}

\begin{table*}[t]
\caption{Performance evaluation with different encoders in Supervised ReID on Market-1501, DukeMTMC-ReID and MSMT-17. Top-left corner shows \textit{Baseline} model trained with real images only, and \textit{ModifiedCamStyle} trained with real images and the full set of CamStyle augmentation, performance shown in the format of Rank@1/mAP(\# of images). '\# of images' indicates the number of generated images used for augmentation.}\label{table:market}
\resizebox{0.92\textwidth}{!}{
\begin{tabular}{|c|c||cc|cc|cc|cc|} 
\hline
\multicolumn{2}{|c||}{\multirow{4}{*}{\begin{tabular}[c]{@{}c@{}}Market-1501 \\ Baseline: 95.1\% / 87.8\% (12936)\\ ModifiedCamStyle: 95.5\% / 87.9\% (64680) \end{tabular}}} & \multicolumn{8}{c|}{$E_d$ } \\ 
\cline{3-10}
\multicolumn{2}{|c||}{} & \multicolumn{2}{c|}{random-initilized} & \multicolumn{2}{c|}{ISGAN-unrelated} & \multicolumn{2}{c|}{ReID-Duke} & \multicolumn{2}{c|}{ImageNet} \\ 
\cline{3-10}
\multicolumn{2}{|c||}{} & Rank@1 & mAP & Rank@1 & mAP & Rank@1 & mAP & Rank@1 & mAP \\
\multicolumn{2}{|c||}{} & \multicolumn{2}{c|}{(\# of images)} & \multicolumn{2}{c|}{(\# of images)} & \multicolumn{2}{c|}{(\# of images)} & \multicolumn{2}{c|}{(\# of images)}  \\ 
\hline
\hline
\multirow{4}{*}{$E_c$ } & \multirow{2}{*}{ISGAN-related} & 95.4\% & 88.2\% & 95.3\% & 88.5\% & 95.6\% & 88.6\% & \bf{96.2\%} & \bf{89.2\%}                      \\ & & \multicolumn{2}{c|}{(18900)} & \multicolumn{2}{c|}{(17733)} & \multicolumn{2}{c|}{(24842)} & \multicolumn{2}{c|}{(17835)} \\ 
\cline{2-10}
& \multirow{2}{*}{ReID-Market} & 95.3\% & 87.9\% & 95.4\% & 88.1\% & 95.4\% & 88.2\% & 95.7\% & 88.7\% \\
& & \multicolumn{2}{c|}{(17988)} & \multicolumn{2}{c|}{(19848)} & \multicolumn{2}{c|}{(17273)} & \multicolumn{2}{c|}{(12771)} \\
\hline
\hline
\hline
\multicolumn{2}{|c||}{\multirow{4}{*}{\begin{tabular}[c]{@{}c@{}}DukeMTMC-ReID\\ Baseline: 89.6\% / 79.1\% (16522)\\ ModifiedCamStyle: 88.4\% / 78.2\% (115654) \end{tabular}}} & \multicolumn{8}{c|}{$E_d$ } \\ 
\cline{3-10}
\multicolumn{2}{|c||}{} & \multicolumn{2}{c|}{random-initilized} & \multicolumn{2}{c|}{ISGAN-unrelated} & \multicolumn{2}{c|}{ReID-Market} & \multicolumn{2}{c|}{ImageNet} \\ 
\cline{3-10}
\multicolumn{2}{|c||}{} & Rank@1 & mAP & Rank@1 & mAP & Rank@1 & mAP & Rank@1 & mAP \\
\multicolumn{2}{|c||}{} & \multicolumn{2}{c|}{(\# of images)} & \multicolumn{2}{c|}{(\# of images)} & \multicolumn{2}{c|}{(\# of images)} & \multicolumn{2}{c|}{(\# of images)}  \\ 
\hline
\hline
\multirow{4}{*}{$E_c$ } & \multirow{2}{*}{ISGAN-related} & 89.0\% & 78.5\% & 89.3\% & 78.8\% & 89.8\% & 79.2\% & \bf{91.0\%} & \bf{79.9\%}                      \\ & & \multicolumn{2}{c|}{(29500)} & \multicolumn{2}{c|}{(28628)} & \multicolumn{2}{c|}{(35849)} & \multicolumn{2}{c|}{(23108)} \\ 
\cline{2-10}
& \multirow{2}{*}{ReID-Duke} & 88.5\% & 77.9\% & 89.1\% & 78.2\% & 89.4\% & 78.9\% & 89.7\% & 79.0\% \\
& & \multicolumn{2}{c|}{(34208)} & \multicolumn{2}{c|}{(33767)} & \multicolumn{2}{c|}{(32384)} & \multicolumn{2}{c|}{(22439)} \\
\hline
\hline
\hline
\multicolumn{2}{|c||}{\multirow{4}{*}{\begin{tabular}[c]{@{}c@{}}MSMT-17\\ Baseline: 81.1\% / 56.5\% (32621)\\ ModifiedCamStyle: 81.0\% / 56.2\% (195726) \end{tabular}}} & \multicolumn{8}{c|}{$E_d$ } \\ 
\cline{3-10}
\multicolumn{2}{|c||}{} & \multicolumn{2}{c|}{random-initilized} & \multicolumn{2}{c|}{ISGAN-unrelated} & \multicolumn{2}{c|}{ReID-Market} & \multicolumn{2}{c|}{ImageNet} \\ 
\cline{3-10}
\multicolumn{2}{|c||}{} & Rank@1 & mAP & Rank@1 & mAP & Rank@1 & mAP & Rank@1 & mAP \\
\multicolumn{2}{|c||}{} & \multicolumn{2}{c|}{(\# of images)} & \multicolumn{2}{c|}{(\# of images)} & \multicolumn{2}{c|}{(\# of images)} & \multicolumn{2}{c|}{(\# of images)}  \\ 
\hline
\hline
\multirow{4}{*}{$E_c$ } & \multirow{2}{*}{ISGAN-related} & 81.3\% & 56.3\% & 81.3\% & 56.4\% & 81.5\% & 56.8\% & \bf{81.9\%} & \bf{57.2\%}                      \\ & & \multicolumn{2}{c|}{(44646)} & \multicolumn{2}{c|}{(45144)} & \multicolumn{2}{c|}{(51965)} & \multicolumn{2}{c|}{(45382)} \\ 
\cline{2-10}
& \multirow{2}{*}{ReID-MSMT} & 81.0\% & 56.2\% & 80.9\% & 55.9\% & 81.1\% & 56.4\% & 81.7\% & 56.6\% \\
& & \multicolumn{2}{c|}{(44356)} & \multicolumn{2}{c|}{(45052)} & \multicolumn{2}{c|}{(49981)} & \multicolumn{2}{c|}{(43256)} \\
\hline
\end{tabular}}
\end{table*}
\section{Experiments}
\label{sec:experiments}

\subsection{Experiment Settings}
\label{ssec:experimentsetting}

\noindent\textbf{Datasets and Evaluation Settings} We evaluate our method on three representative ReID datasets, Market-1501~\cite{zheng2015scalable}, DukeMTMC-reID~\cite{zheng2017unlabeled} and MSMT-17~\cite{wei2018person}. 
Cumulative Matching Characteristic (CMC) at rank-1 and mAP~\cite{zheng2015scalable} are used as the evaluation metrics.

\noindent\textbf{GAN Model for Augmentation} CamStyle~\cite{zhong2018camera} has proven its ability to generate person images across different camera style, but it has two weaknesses. First it is based on the CycleGAN~\cite{zhu2017unpaired}, which requires to train a translation model for each camera pair. For a N-camera dataset, it would be too time-consuming to train $N(N-1)$ models. To solve this problem, we import StarGAN~\cite{choi2018stargan} into the CamStyle~\cite{zhong2018camera} which allows us to train multi-camera image-image translation with a single model. 

Second, the original backbone of CamStyle model only contains several layers, which is hard to take full advantages of CamStyle.
So we modify the backbone carefully. Specifically, the modified generator consists of four down-sampling blocks, four intermediate blocks and four up-sampling blocks. Instance normalization (IN)~\cite{chen2019instance} and adaptive instance normalization (AdaIN)~\cite{huang2017arbitrary} are used for down-sampling and up-sampling respectively. Discriminator is a multi-task discriminator~\cite{mescheder2018training}, which contains $L$ linear output branches, where $L$ indicates the number of cameras. The discriminator contains six pre-activation residual blocks with leaky ReLU~\cite{maas2013rectifier}. Finally, we use Adam optimizer~\cite{kingma2014adam} with $\beta_{1}=0.5$ and $\beta{2}=0.999$. The learning rates for $G$ and $D$ are set to $10^{-4}$.
The results of the original CamStyle (\textit{OriginalCamstyle}) and our modified CamStyle (\textit{ModifiedCamStyle}) are shown in Table.~\ref{table:SupSOTA}. It is obvious that our modified CamStyle provides a much better performance than the original CamStyle. And the augmented images of the original CamStyle is even worse than the Baseline.

\noindent\textbf{ReID Baseline Model} Our method can be applied to many ReID models, this paper uses reid-strongbaselne~\cite{luo2019bag} as an example baseline to verify the effectiveness of our approach. We train baseline model with 256x128 input images, which is processed by random cropping~\cite{krizhevsky2017imagenet}, random horizontal flipping~\cite{simonyan2014very} and random erasing~\cite{zhong2020random}. We perform 60 epochs training on the model, using $B = P \times K = 6\times 9$ as mini-batch, where $P=6$ indicates the number of person-ids and $K=9$ indicates the number of real images per person-id from training dataset. 


\subsection{Comparison with State of the arts}
\label{ssec:sota}

The comparison with state-of-the-arts is shown in Table.\ref{table:SupSOTA}. \textit{Baseline} is our implementation based on reid-strongbaseline~\cite{luo2019bag}, where only the real data are used for supervised ReID training. The original CamStyle and the StarGAN-based CamStyle are employed to generate images based on the real data which are considered as full augmentation, and denoted as \textit{OriginalCamStyle} and \textit{ModifiedCamStyle} respectively. \textit{Ours} shows the results of our method, which samples the generated images of StarGAN-based CamStyle with id-consistency and diversity constraints before adding into the training. 
With a state-of-the-art baseline model, the \textit{ModifiedCamStyle} only brings a slight improvement on some datasets. Comparing \textit{Ours} with \textit{ModifiedCamStyle}, the performance is improved largely, which suggest that NOT every generated image benefits the ReID training. Eliminating the augmentations that do not satisfy our constraints could further improve the effectiveness of data augmentation.

Figure.~\ref{fig.3} gives the t-SNE~\cite{maaten2008visualizing} visualization of feature distribution on Market-1501, circle dots denote real images and triangles denote generated images. Although full augmentations can fill in gaps between real images and extend class boundaries, it also brings in a large number of noisy or duplicate samples, as shown in Figure. ~\ref{fig.3}(a), which have negative impact on ReID training. Figure.~\ref{fig.3}(b) shows the feature distribution after the generated samples being filtered based on proposed constraints. The removal of noise samples and duplicate samples help defining more reasonable distribution of training data, making it easier for ReID model to learn a robust yet discriminative model with better performance.

\subsection{Ablation Study}
\label{ssec:ablation}

In this section, we firstly explore different choices of encoders to build the consistency and diversity spaces. Then we conduct ablation studies to evaluate each component in our method.

\noindent\textbf{Consistency Encoders} Table.\ref{table:market} shows the results of different encoders on Market-1501, DukeMTMC-ReID and MSMT-17. Before comparison, on the left top, $Baseline$ represents the result without any GAN augmentation and $ModifiedCamStyle$ represents using full set of CamStyle augmentation. The number of augmented fake images is listed in the brackets behind the performance. It is worth noting that MSMT-17 has 15 cameras and there will be 247380 images, which is a large number. In order to save disk space and training time, we randomly select six cameras to do camstyle.

The results of different consistency encoders are listed in different rows.
Intuitively, a state-of-the-art ReID model itself should be powerful enough to extract id-consistency information. Therefore, besides using the ISGAN id-related model to serve as $E_c$, we also train a state-of-the-art ReID model\cite{luo2019bag} as an alternative choice to build the consistency space. ISGAN id-related model and the state-of-the-art ReID model are denoted as $ISGAN\!-\!related$ and $ReID\!-\!X$ respectively, where $X$ is the dataset name.

Taking Market-1501 as an example, comparing the top row and bottom row of Table \ref{table:market}, it can be observed that the consistency space built through ISGAN id-related branch outperforms the space built through the ReID model, even though the ReID model has already been carefully trained. It is expected that the generated images should preserve as much identity information as possible. ISGAN employs additional supervision and a shuffle strategy to enhance the ability of identity consistent representation. This conclusion is also consistent on DukeMTMC-ReID and MSMT-17.

\begin{figure}[t]
\begin{center}
\includegraphics[width=0.44\textwidth]{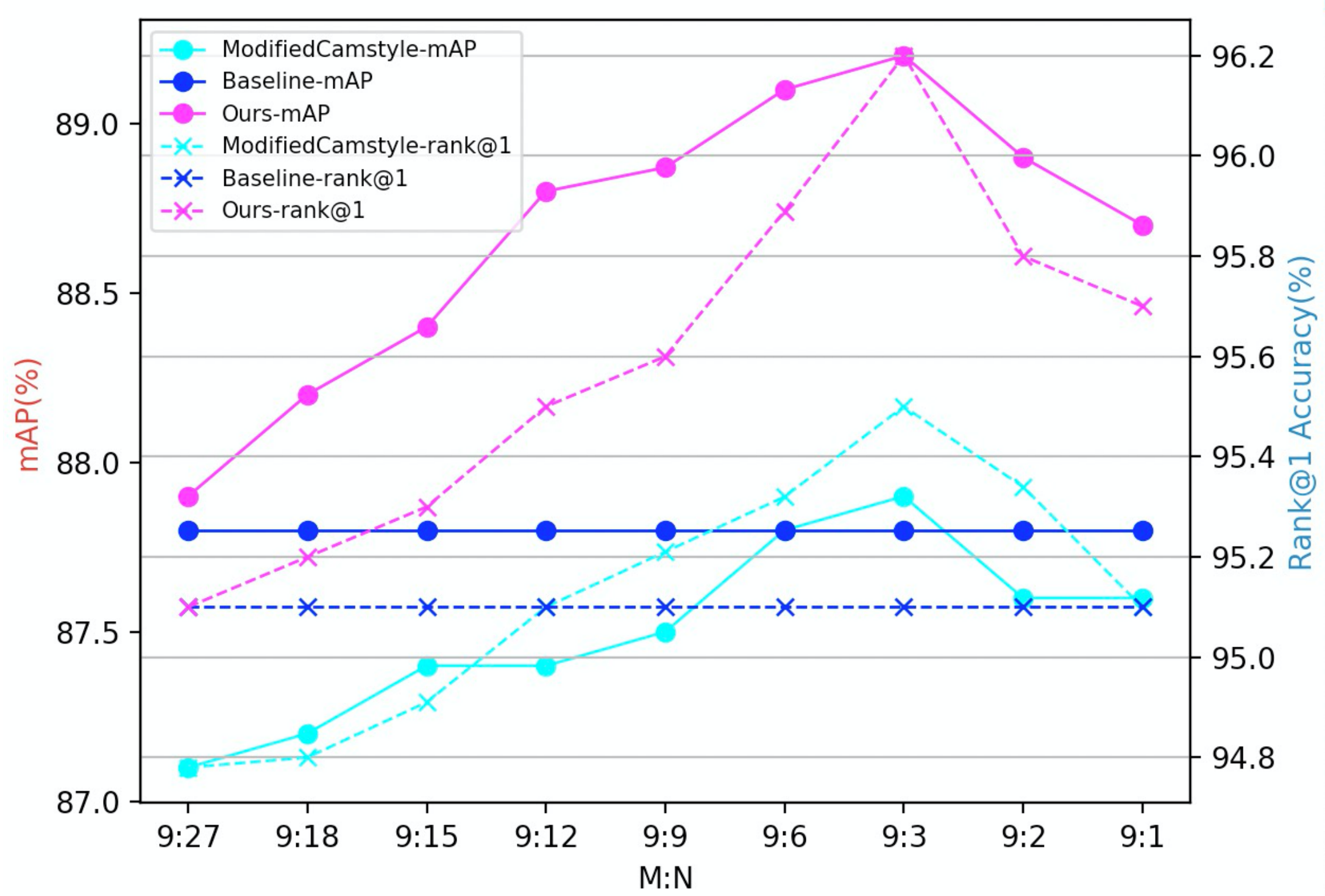}
\end{center}
   \caption{Evaluation with different ratio of real data and sampled augmentations ($M:N$) in a mini-batch on Market-1501.}
   \Description{Evaluation with different ratio of real data and sampled augmentations in a mini-batch on Market-1501. Left y-axis is the performance of mAP, right y-axis is mAP. It can be seen, Ours with different ratio consistently improve over baseline, while ModifiedCamStyle achieves lower performance than baseline when using more fake data than real data.}
\label{fig:marketratio}
\end{figure}

\begin{table}
\caption{Performance Evaluation on Market-1501 using different sets. '\# of images' indicates the number of generated images used for augmentation.} 
\label{table:marketsample}
\resizebox{0.44\textwidth}{!}{
\begin{tabular}{|p{2.2cm}<{\centering}|p{1.6cm}<{\centering}|p{1cm}<{\centering}|p{1cm}<{\centering}|p{1cm}<{\centering}|}
\hline
Method & \# of Images & Rank@1 & mAP \\
\hline
\hline
Baseline & - & 95.1 & 87.8 \\
ModifiedCamStyle & 64680 & 95.5 & 87.9 \\
Random & 17835 & 94.9 & 87.6 \\ 
Consistency & 32340 & 95.4 & 88.3 \\
Diversity & 32340 & 95.7 & 88.9 \\
Ignored & 46845 & 94.2 & 87.1 \\
Ours & 20386 & 95.6 & 88.8 \\
Ours $\bf{w/}$ lof & 17835 & \textbf{96.2} & \textbf{89.2} \\
\hline
\end{tabular}}
\end{table}
\begin{table}
\caption{Performance Evaluation on Market-1501 using different threshold. \textit{$T_c$} and \textit{$T_d$} are the thresholds for consistency space and diversity space respectively. '\# of images' indicates the number of generated images used for augmentation.}
\label{table:marketthreshold}
\resizebox{0.44\textwidth}{!}{
\begin{tabular}{|p{1cm}<{\centering}|p{1cm}<{\centering}|p{1.8cm}<{\centering}|p{1cm}<{\centering}|p{1cm}<{\centering}|}
\hline
$T_c$ & $T_d$ & \# of Images & Rank@1 & mAP \\
\hline
\hline
\multicolumn{2}{|c|}{Baseline} & - & 95.1 & 87.8 \\
\hline
mean & mean & 16996 & 95.6 & 88.9 \\ 
mean & median & 17990 & 95.8 & 89.1 \\
median & mean & 15162 & 95.6 & 88.8 \\
median & median & 17835 & \textbf{96.2} & \textbf{89.2} \\
\hline
\end{tabular}}
\end{table}

\begin{table}
\caption{Performance Evaluation on Market-1501 using different loss functions. ID: Cross-Entropy Loss, Tri: Triplet Loss.}
\label{table:marketloss}
\resizebox{0.44\textwidth}{!}{
\begin{tabular}{|p{1.8cm}<{\centering}|p{1.8cm}<{\centering}|p{1cm}<{\centering}|p{1cm}<{\centering}|p{1cm}<{\centering}|}
\hline
Training data & $L_R$ & $L_F$ & Rank@1 & mAP \\
\hline
\hline
Real & ID+Tri & None & 95.1 & 87.8 \\ 
Real+Fake & ID+Tri & ID & \textbf{96.2} & \textbf{89.2} \\
Real+Fake & ID+Tri & Tri & 93.2 & 86.1 \\
Real+Fake & ID+Tri & ID+Tri & 94.3 & 86.8 \\
\hline
\end{tabular}}
\end{table}

\noindent\textbf{Diversity Encoders} Four different choices of the diversity feature space encoders are explored in columns in Table.\ref{table:market}: (1) a model with random initialized parameters, denoted as \textit{random-initialized}; (2) a model from ISGAN id-unrelated branch, denoted as \textit{ISGAN-unrelated}; (3) a well-trained ReID model on dataset in different domain, denoted as \textit{ReID-X} with \textit{X} representing the domain different from consistent space; (4) the proposed model pretrained on ImageNet, denoted as \textit{ImageNet}. We can see that \textit{ImageNet} outperforms all other choices of diversity encoders. As we argued in diversity constraint, the generated image should contain as much diversity changes as possible. The ImageNet pretrained model generates a feature space with rich texture, illumination and other representations. \textit{random-initialized} and \textit{ISGAN-unrelated} produce the lowest performances, because the space built by these two models is either a random space or a Gaussian noise space, which captures little information about image diversity.  The \textit{ReID-X} encoder outperforms \textit{random-initialized} but is inferior to \textit{ImageNet}. This is because the ReID diversity encoder is trained on the data from another domain, which could bring some variance.
But it still contains a certain degree of the consistency, which leads to some conflicts with the consistency space during sampling. For instance, some ''best augmentation sample'' could be eliminated in diversity space since they are with the same identity and with insufficient distance from the id-center in diversity space. Consequently, the sampling result of ReID diversity encoder is not comparable with the result of \textit{ImageNet} encoder.

\begin{table*}[t]
\caption{Comparisons with the baseline model and state-of-the-art UDA ReID methods between Market-1501 and DukeMTMC-ReID datasets (\%). The best and second-best results are presented in \textbf{bold} and \underline{underline}.}
\label{table:UDASOTA}
\resizebox{0.91\textwidth}{!}{
\begin{tabular}{|l|p{1.8cm}<{\centering}|p{1cm}<{\centering}|p{1cm}<{\centering}|p{1cm}<{\centering}|p{1.2cm}<{\centering}|p{1cm}<{\centering}|p{1cm}<{\centering}|p{1cm}<{\centering}|p{1.2cm}<{\centering}|}
\hline
\multicolumn{2}{|c|}{\multirow{2}{*}{Method}} & \multicolumn{4}{c|}{Market-1501 $\to$ DukeMTMC-reID} & \multicolumn{4}{c|}{DukeMTMC-reID $\to$ Market-1501} \\
\cline{3-10}
\multicolumn{2}{|c|}{} & mAP & Rank@1 & Rank@5 & Rank@10 & mAP  & Rank@1  & Rank@5 & Rank@10 \\
\hline
\hline
PT-GAN \cite{wei2018person} & CVPR2018 & - & 27.4 & - & 50.7 & - & 38.6 & - & 66.1 \\  %
SPGAN-LMP \cite{deng2018image} & CVPR2018 & 26.4 & 46.9 & 62.6 & 68.5 & 26.9 & 58.1 & 76.0 & 82.7 \\ %
ECN \cite{zhong2019invariance} & CVPR2019 & 40.4 & 63.3 & 75.8 & 80.4 & 43.0 & 75.1 & 87.6 & 81.6 \\ %
Theory \cite{song2020unsupervised} & PR2020 & 49.0 & 68.4 & 80.1 & 83.5 & 53.7 & 75.8 & 89.5 & 93.2 \\
SSG \cite{fu2019self} & CVPR2019 & 53.4 & 73.0 & 80.6 & 83.2 & 58.3 & 80.0 & 90.0 & 92.4 \\
AD-Cluster \cite{zhai2020ad} & CVPR2020 & 54.1 & 72.6 & 82.5 & 85.5 & 68.3 & 86.7 & 94.4 & 96.5 \\
MMCL \cite{wang2020unsupervised} & CVPR2020 & 51.4 & 72.4 & 82.9 & 85.0 & 60.4 & 84.4 & 92.8 & 95.0 \\
JVTC \cite{li2020joint} & ECCV2020 & 56.2 & 75.0 & 85.1 & 88.2 & 61.1 & 83.8 & 93.0 & 95.2 \\
NMRT \cite{zhao2020unsupervised} & ECCV2020 & 62.2 & 77.8 & 86.9 & 89.5 & 71.7 & 87.8 & 94.6 & 96.5 \\
MMT \cite{ge2020mutual} & ICLR2020 & {\bf 68.7} & \bf{81.8} & {\bf 91.2} & {\bf 93.4} & 74.5 & \underline{91.1} & \bf{96.5} & \bf{98.2} \\
\hline
Baseline & - & 56.1 & 74.1 & 84.1 & 87.3 & 62.8 & 84.7 & 93.4 & 95.6 \\
ModifiedCamStyle & - & 58.7 & 76.0 & 85.7 & 89.5 & \underline{74.9} & 90.8 & 95.6 & 97.3 \\
{\bf Ours} & - & {\underline {64.4}} & \underline{79.4} & {\underline {88.6}} & {\underline {91.0}} & {\bf 78.6} & \bf{91.3} & \underline{96.3} & \underline{97.7} \\
\hline
\end{tabular}}
\end{table*}


\noindent\textbf{Sampling Method Analysis} Table.~\ref{table:marketsample} shows the effectiveness of the proposed sampling method. $ModifiedCamStyle$ denotes using full fake images for augmentation. $Random$ denotes randomly selecting the same number of fake images as our proposed method. $Consistency$ and $Diversity$ denotes using only id-consistency constraint and diversity constraint, respectively. $Ignore$ denotes selecting samples that are ignored by our proposed method. We can see that $Random$ and $Ignore$ has the lower performance while $Consistency$ and $Diversity$ consistently gets higher performance than $Baseline$. $Ours$ uses both id-consistency and diversity constraint to do sampling and gets much higher performance. Considering that generated images which hold a high density do not bring in much diversity information and need to be dropped, $Ours$ $w/$ $lof$ uses LOF monitor to further improve the quality of augmentation data set, which also partly handles the data imbalance problem and achieves the best performance. The best performance of $Ours$ $w/$ $lof$ shows the effectiveness of the monitor and the monitor can be regarded as a beneficial complement to our proposed constraints.

\noindent\textbf{Threshold Analysis} Threshold $T_c$ and $T_d$ in Equ.\ref{equ:sfinal} are important parameters in our sampling process. We explore the median value and the mean value of the distances from the center of all pictures in each category, as our thresholds. The results are shown in Table~\ref{table:marketthreshold}. The performance of ReID model varies with the setting of threshold, but they all consistently higher than baseline, which shows the effectiveness of consistency constraints and diversity constraints. Meanwhile, using median as threshold get highest performance, and it is used as our default setting through all the experiments. 

\noindent\textbf{Data Ratio Analysis} The ratio of $\frac{M}{N}$ is critical for ReID model training, where $M$ and $N$ indicate the number of real and fake samples in mini-batch. Figure.~\ref{fig:marketratio} shows results on different ratio. The x-axis shows the different $M$ and $N$ used for training. Our proposed method with different $\frac{M}{N}$ consistently improve over baseline, while \textit{ModifiedCamStyle} achieves lower performance than baseline when using more fake data than real data. We argue that \textit{ModifiedCamStyle} contains a large number of fake images that do not meet id-consistency and diversity constraints, and are not suitable for ReID training. \textit{Ours} filters out those fake images and achieve the best performance with the data ratio $M:N=9:3$.

\noindent\textbf{Loss Function Analysis} Table.~\ref{table:marketloss} gives results of using different loss functions on sampled fake images. Recently works have proved that using ID loss and triplet loss together to train ReID model is beneficial. However, ReID performance drops a lot when we apply either two losses or only triplet loss on sampled augmentations. A possible explanation is that there still exists noise in fake images even after our sampling. The triplet loss is more sensitive to these noises than the classification loss, resulting in worse result.

\subsection{Experiments on UDA ReID Task}

Besides the work on the Supervised ReID task, we also conduct experiments on Unsupervised Domain Adaptive(UDA) ReID to further show the effectiveness of our method. The comparison is shown in Table.\ref{table:UDASOTA}. \textit{Baseline} is our implementation based on \cite{song2020unsupervised}, where only the real data in target domain are used for UDA ReID training. StarGAN-based CamStyle is employed to generate images based on the target-domain real data, which is considered as full augmentation and denoted as \textit{ModifiedCamStyle}. \textit{Ours} shows the results of our method, which samples the generated images of StarGAN-based CamStyle with the proposed two constraints before adding into the training. 

The \textit{ModifiedCamStyle} increases the \textit{Baseline} performance of UDA ReID without bells and whistles. Comparing \textit{Ours} with \textit{ModifiedCamStyle}, the performance is further improved, which suggest that NOT every generated image benefits the ReID training. Eliminating the augmentations that do not satisfy our constraints could further improve the effectiveness of data augmentation.
More interesting, we can find that a low baseline (\textit{i.e.} 84.7\%/62.8\%) can be improved to a state-of-the-art result (\textit{i.e.} 91.3\%/78.6\%) with the help of our method.

\section{Conclusion}

In this paper, we claim that not every GAN generated sample is beneficial for ReID Training. A good augmentation should satisfy both identity consistency constraint and diversity constraint. We propose to project person image into consistency and diversity feature spaces and select good augmentations through a simple distance metric sampling method. Experiments on several benchmarks demonstrate that the constraints on both consistency and diversity are necessary for GAN based augmentation.
We believe that the consistency and diversity spaces could be further explored with other well-designed encoders. Besides, a more complicated sampling method which exploits the interaction between samples or performs a combinatorial optimization, could yield better sub set of augmentation. We leave these as our future work.

\bibliographystyle{ACM-Reference-Format}
\bibliography{sample-base}
\end{document}